\documentclass[letterpaper, 10 pt, conference]{ieeeconf}  

\IEEEoverridecommandlockouts                              

\usepackage{balance}
\usepackage{times}
\usepackage{epsfig}
\usepackage{graphicx}
\usepackage{amsmath}
\usepackage{amssymb}
\usepackage{subfigure}
\usepackage[ruled]{algorithm}
\usepackage{algpseudocode}
\usepackage{ctable}
\usepackage{caption}
\usepackage[export]{adjustbox}

\makeatletter
\let\NAT@parse\undefined
\makeatother
\usepackage[numbers]{natbib}

\usepackage{irap_acronyms}
\usepackage{irap_math}
\usepackage{irap_SIunits}
\usepackage{irap_misc}
\usepackage{multirow}

\usepackage{soul,color}
\usepackage{verbatim} 

\usepackage[symbol]{footmisc}

\usepackage{xcolor}

\usepackage[symbol]{footmisc}

\title{\LARGE \bf
  DeepLO: Geometry-Aware Deep LiDAR Odometry
}

\author{Younggun Cho${}^{1}$, Giseop Kim${}^{1}$ and Ayoung Kim${}^{1*}$
\thanks{$^{1}$Y. Cho, G. Kim and A. Kim are with the Department of Civil and
        Environmental Engineering, KAIST, Daejeon, S. Korea
        {\tt\small [yg.cho, paulgkim, ayoungk]@kaist.ac.kr}}%
\thanks{This work is supported by the Korea MOLIT (19CTAP-C142170-02).}%
}

\begin{document}

\maketitle
\thispagestyle{empty}
\pagestyle{empty}


\begin{abstract}

Recently, learning-based ego-motion estimation approaches have drawn strong
interest from studies mostly focusing on visual perception. These groundbreaking
works focus on unsupervised learning for odometry estimation but mostly for
visual sensors. Compared to images, a learning-based approach using \ac{LiDAR}
has been reported in a few studies where, most often, a supervised learning
framework is proposed. In this paper, we propose a novel approach to
geometry-aware deep LiDAR odometry trainable via both supervised and
unsupervised frameworks. We incorporate the \ac{ICP} algorithm into a
deep-learning framework and show the reliability of the proposed pipeline.  We
provide two loss functions that allow switching between \textit{supervised} and
\textit{unsupervised} learning depending on the ground-truth validity in the
training phase. An evaluation using the KITTI and Oxford RobotCar dataset
demonstrates the prominent performance and efficiency of the proposed method
when achieving pose accuracy. The overall
algorithm is presented in https://youtu.be/Y2s08dv-Mq0.

\end{abstract}


\section{Introduction}

Odometry (ego-motion) estimation is a core module in \ac{SLAM} which presents
various applications to an autonomous robot \cite{cadena2016past} and 3D mapping
\cite{serafin2015nicp,maddern20171}. So far, most odometry modules have been
focused on model-based using cameras \cite{mur2015orb, forster2014svo,
engel2018direct} and \ac{LiDAR} \cite{bosse2012zebedee, Zhang-RSS-14,
Behley-RSS-18}. For example, visual-LiDAR odometry and mapping (V-LOAM) records
the first place in the KITTI odometry benchmark and has shown remarkable
accuracy. Despite their superior performances, model-based methods are exposed
to challenges such as vulnerability to environmental disturbance and parameter
selection. Therefore, recent studies have started to examine learning-based
methods mostly for visual odometry in both a supervised \cite{wang2017deepvo,
zhou2018deeptam} and an unsupervised \cite{zhou2017unsupervised, li2018undeepvo}
manner.

Similar to vision, some effort toward learning-based odometry using a
\textit{range sensor (e.g. LiDAR)} has been initiated. However, the major
challenge is to handle a dense point cloud by feeding it into a deep neural
network, and several recent studies have focused on feeding the point cloud directly to the
network \cite{qi2017pointnet, lee2018pointnetvlad}, albeit for object-sized point
cloud data. As an example for odometry, learning-based approaches for point clouds
were presented in \cite{nicolai2016deep, velas2018cnn} where the authors
relied on a supervised method requiring the ground-truth with labeled sequences.
Unlike these previous approaches, we examine an unsupervised manner for
deep LiDAR odometry in order to achieve scalability and flexibility in the
training phase.

\begin{figure}[!t]
    \centering
    \includegraphics[width=0.9\columnwidth]{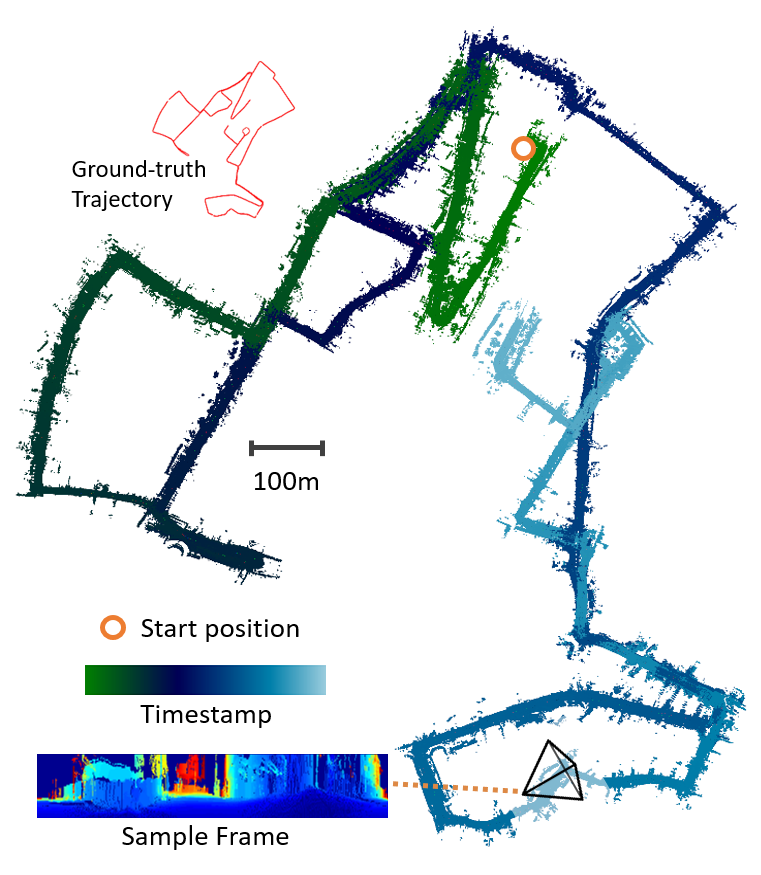}%

    \caption{A point cloud map using learned LiDAR odometry. The figure shows
    long sequences of the Oxford RobotCar dataset \cite{maddern20171}. The
    oragne circle indicates the start position and point clouds are colored with
    respect to timestamps (mission time). A sample LiDAR frame is also
    presented.}

    \label{fig:thumbnail_map}
    \vspace{-4mm}
\end{figure}

In this paper, we propose unsupervised deep LiDAR odometry, called
\textit{DeepLO}. For efficiency, we feed a rendered vertex map and a normal map
into a network and regress a 6D relative pose between two frames, while the
NICP-like loss \cite{serafin2015nicp} is calculated using the aforementioned
representations; thus, an overall training pipeline is conducted in an
unsupervised manner. \figref{fig:thumbnail_map} shows the learned trajectory of
Oxford Robotcat dataset \cite{maddern20171} with an unsupervised manner. This
figure indicates that our method successfully captures the relative motion of a
long sequence trajectory (\unit{10}{km}) without ground-truth. To the best of
our knowledge, our work is the first unsupervised learning-based odometry for a
range sensor.

Our contributions are:

\begin{itemize}

  \item We propose a general pipeline for deep-learning-based LiDAR odometry that can
  be trained in both a supervised and an unsupervised manner.

  \item For efficient unsupervised training and inference, we use a vertex and
  normal map as inputs and use them on loss calculation. By doing so, the
  time-consuming labeling procedure is alleviated in an unsupervised fashion.
  These summarized representations can be exploited in both the training and
  inference stages.



  \item The proposed learning system can generally be used for a \ac{LiDAR}
  point cloud (submap) regardless of the hardware type or configuration (e.g.,
  the 3D surrounding by KITTI dataset and the 2D push-broom of the Oxford
  RobotCar dataset).

\end{itemize}

The remaineder of the paper is composed as follows. In \secref{related_works}, recent
works on model and learning-based odometry estimation methods are discussed.
In \secref{proposed_method}, we introduce the architecture and loss functions
of our learning-based LiDAR odometry. The performance of our method is compared
with other state-of-the-art methods in \secref{experimental_results}. Conclusions and ideas for further work are shared in \secref{conclusion}.

\section{Related Works}
\label{related_works}

In this section, we provide a summary of existing model- and learning-based
methods for odometry estimation.


\subsection{Model-based Odometry Estimation for a Range Sensor}
\label{model_odom}

For range sensors such as an RGB-D camera and LiDAR, many model-based methods
\cite{Zhang-RSS-14, whelan2016elasticfusion, Behley-RSS-18} including the odometry
module, have been proposed which minimize the error between two consecutive frames
or a frame and a map using the \ac{ICP} \cite{icp1992, rusinkiewicz2001efficient,
Segal-RSS-09, serafin2015nicp}. LOAM extracts edge and
planar features for matching and run two parallel modules of different
frequencies for fast and accurate ego-motion estimation \cite{Zhang-RSS-14}. Recently,
\citeauthor{Behley-RSS-18} proposed a surfel-based mapping
method, called SuMa, for 3D laser range data \cite{Behley-RSS-18}. The SuMa representation is considered to be efficient and accurate for dense mapping such as ElasticFusion \cite{whelan2016elasticfusion} using an RGB-D camera. This type of
research \cite{whelan2016elasticfusion, Behley-RSS-18} also minimizes the error
between the current frame and the rendered view of a map using \ac{ICP}. By
leveraging rendered-image coordinate-parameterized normal and vertex
information, SuMa can almost employ points for \ac{ICP} (unlike feature-based
methods such as LOAM) while minimizing a loss of a piece of original information
(e.g., from filtering, rasterization, or taking features).

\subsection{Learning-based Odometry Estimation}
\label{learning_odom}

\subsubsection{Visual Sensor} Recently a number of learning-based visual
odometry methods have been developed. \citet{wang2017deepvo} proposed a
supervised visual odometry network using a recurrent network structure, while \citet{zhou2018deeptam} introduced a method of configuring a tracking and
mapping process in a network structure. However, since it is challenging to
construct a lot of ground-truth data to train a network, many unsupervised
learning methods which leverage photo-consistency have recently been introduced.
\citet{zhou2017unsupervised} reported that the metric depth could be learned from
monocular sequences by warping the consecutive images. \citet{li2018undeepvo}
proposed the self-supervised learning of depth and ego-motion through spatial and
temporal stereo consistency. Recently, \cite{luo2018every} proposed the joint
learning of the depth, odometry, and optical flow of consecutive scenes with
consideration of the outlier (dynamic) pixels for robustness. In addition,
\cite{zhu2018robustness} introduced a hybrid pipeline for robust ego-motion
estimation. This method learns disparity and depth via deep networks and
predicts relative motion with model-based \ac{RANSAC} outlier rejection.

\subsubsection{Range Sensor} Unlike the aforementioned methods for visual
sensors, there are few learning-based methods for range sensors because the
range sensor data (e.g., 3D point cloud) is sparse and irregular; this fact
makes it difficult to directly employ conventional modules such as 2D
convolution and upconvolution due to memory inefficiency issue. Recently, some
methods that  consume irregular point cloud data directly and achieve
permutation invariance have been proposed for object recognition or the
segmentation problem \cite{qi2017pointnet, li2018pointcnn}. However, to the best
of our knowledge, there have not yet been any empirical reports for odometry
estimation that directly leverages point cloud. To avoid the these issues, a few
studies proposed rasterized image-based  learning methods for 3D LiDAR odometry
\cite{nicolai2016deep, velas2018cnn}. However, they lose the original point
cloud information (e.g., the 3D point coordinates as real numbers) via the
rasterization, and their training is performed in a supervised manner, thus they
have low scalablility for the emergently available 3D point cloud data as LiDAR
becomes more popular.


In contrast to the aforementioned methods, we propose a deep \ac{LiDAR} odometry
network in both an unsupervised and a supervised manners. By incorporating
traditional model-based point errors into the deep architecture, we can build
the unsupervised learning pipeline of \ac{LiDAR}-based odometry without ground-truth relative poses.


\section{Proposed Method}
\label{proposed_method}

\begin{figure*}[!t]
    \centering
    \includegraphics[clip, trim=30 40 20 30,
    width=0.95\textwidth]{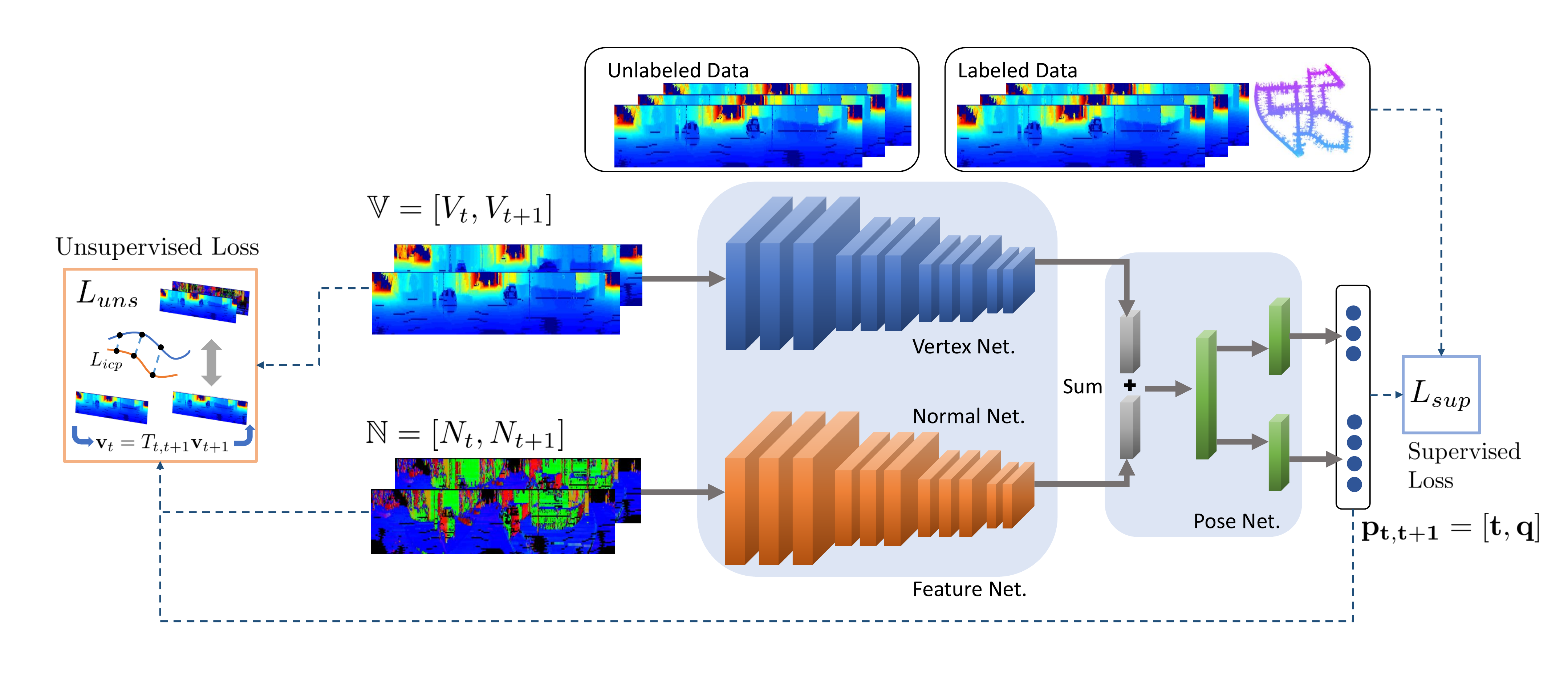}%

    \caption{The proposed network and our unsupervised training scheme. The
    network is composed of \textit{FeatNet} (composed of \textit{VertexNet} and
    \textit{NormalNet}) and \textit{PoseNet}. First, as a common part,
    sequential frames ($F_t$ and $F_{t+1}$) are fed into the feature extractor
    (\textit{FeatNet}) for a compact representation. Subsequently, each frame
    feature is summed as a single vector and forwarded into \textit{PoseNet}
    which predicts the relative motion $\mathbf{p}_{t,t+1}=[\mathbf{t},
    \mathbf{q}]$ of two frames. For the supervised mode when ground-truth motion
    is available, the predicted motion is converted to the Euler form and
    directly compared to the ground-truth ($\mathcal{L}_{sup}$). For the
    unsupervised mode, the output is used to compute the unsupervised loss (ICP
    loss $\mathcal{L}_{icp}$ and FOV loss $\mathcal{L}_{fov}$) without any
    ground-truth motion.}

    \label{fig:network}
    \vspace{-4mm}
\end{figure*}

In this section, we describe the details of our approach. Our system is composed
of feature networks (\textit{FeatNet}) and a pose network (\textit{PoseNet}).
FeatNets extract the feature vectors of consecutive frames and PoseNet estimates
the relative motion of the frames from features. The networks can be trained in
both a supervised and an unsupervised manner. A proper training strategy can be
chosen according to the availability of ground-truth labels. The overall
pipeline for training and  inference is depicted in \figref{fig:network}.

\begin{figure}[!t]
    \centering
    \includegraphics[width=0.9\columnwidth]{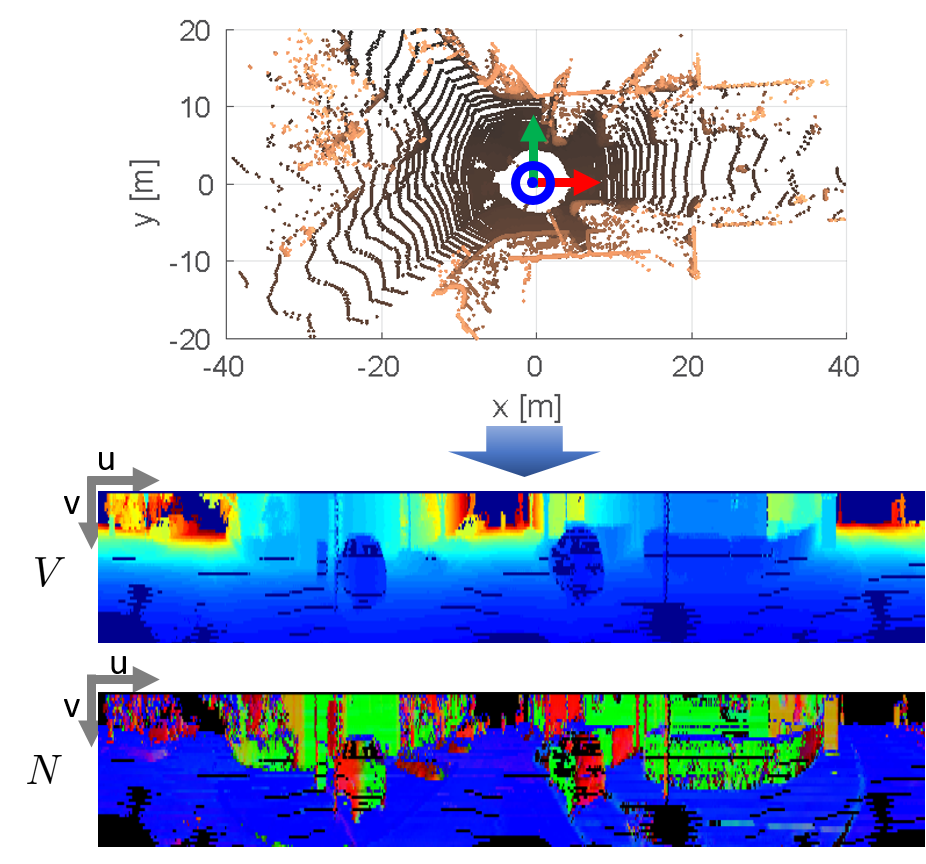}%

    \caption{LiDAR-induced vertex ($V$) and normal ($N$) maps were used as input
    for the network. The first row shows raw point clouds with the local axis in
    the center (RGB represents the XYZ axis). The bottom rows are input vertex
    and normal maps. The vertex map is color-coded with respect to the range
    from the origin for visualization.}

    \label{fig:input_repre}
    \vspace{-4mm}
\end{figure}

\subsection{Input Representation}

Before describing the details, we first explain the input representation given a
\ac{LiDAR} point cloud. To cope with the unordered characteristics of a
\ac{LiDAR} point cloud, we reformulate it using an image
coordinate-parameterized representation which unlike rasterized image (e.g.,
range image in \cite{velas2018cnn}), preserves the 3D point information as real
numbers. We employ projection function $\pi(\cdot) : \mathbb{R}^3 \mapsto
\mathbb{R}^2$ to project the 3D point cloud into 2D image plane on spherical
coordinates. Each 3D point $\mathbf{p} = (p^x, p^y, p^z)$ in a sensor frame is
mapped onto the 2D image plane $(u, v)$ represented as
\begin{equation}
  \binom{u}{v} = \binom{(f_h/2 - \arctan(p^y, 
      p^x))/\delta_h}{(f_{vu}-\arctan(p^z,
d))/\delta_v},
\end{equation}
where depth $d = ({p^x}^2+{p^y}^2)^{1/2}$; $f_h$ and $f_v$ are the horizontal
(azimuth) and vertical (elevation) field-of-view, respectively; vertical
field-of-view $f_v = f_{vu} + f_{vl}$ is composed of upper ($f_{vu}$) and lower
($f_{vl}$) parts. Here, $\delta_u$ and $\delta_v$ are the horizontal and
vertical resolutions for pixel representation. If several 3D points are
projected onto the same pixel coordinates, we choose the nearest point as a
pixel value. We define the mapped representation as vertex map $V$ which has 2D
coordinates $(u,v) \in \mathbb{R}^2$ and 3-channel values $\mathbf{v}=[v^x, v^y,
v^z]$ as a 3D point. \figref{fig:input_repre} shows an example point cloud $P_t$
on timestamp $t$, corresponding vertex map $V$, and normal map $N$ on frame
$F_t=[V_t, N_t]$.

We then assign a normal vector $\mathbf{n}$ of each vertex $\mathbf{v}$ adopting
normal estimation methods in \cite{serafin2015nicp}. The normal vector of each
vertex $\mathbf{v}$ is computed by the nearest vertices in the vertex map.
Because we already built a vertex map,  extensive queries on a kd-tree are not
required. For reliable normal vector estimation, we discard the distance
vertices from the center vertex. In this paper, we set the threshold range as
\unit{50}{cm} and filtered simply by the depth values computed on vertex map
generation.

\begin{figure}[!t]
    \centering
    \includegraphics[width=1\columnwidth]{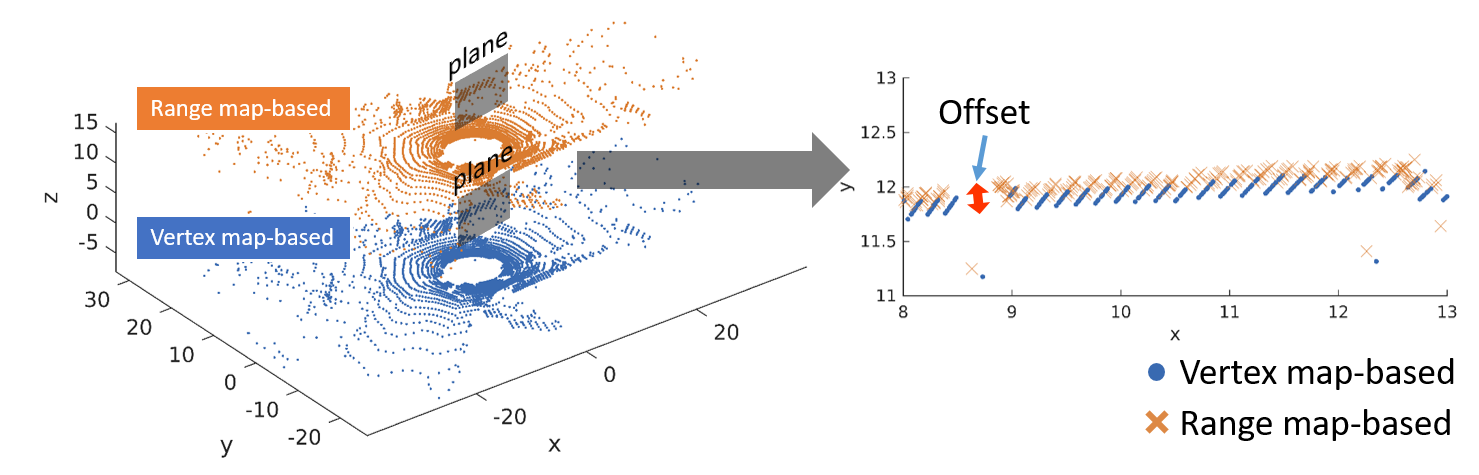}%

    \caption{A comparison between two point cloud representations. The
    reconstructed point cloud from the proposed method (blue) and the range
    map-based method (orange). As shown in the enlarged plot over a plane
    (right), the range map-based representation has discretization errors on the
    points.}

    \label{fig:pcd_comparison}
    \vspace{-4mm}
\end{figure}

To verify the frame representation $F$, we compare the proposed representation
with the existing range-based representation which utilizes a point range and
extra characteristics (e.g., intensity and height) as pixel values. 
\figref{fig:pcd_comparison} shows the comparisons of reconstructed point clouds
of each descriptions: vertex map (blue) and range map (orange) representation.
The left plot is a sample frame of an urban scene (left), and a right plot
represents a top view of a wall (plane). As in the enlarged view of wall, the
range map-based representation \cite{velas2018cnn,nicolai2016deep} has offset of
the reconstructed point clouds due to the angular discretization of the range
image. On the other hand, the vertex map is represented by a raw point cloud,
discretization offsets can be prevented.

\subsection{Proposed Network}

The proposed network is composed of two parts, as shown in \figref{fig:network}:
a vertex network (\textit{VertexNet}), a normal network (\textit{NormalNet}), and the
pose networks (\textit{PoseNet}). \textit{VertexNet} and \textit{NormalNet} use
the vertex and normal maps as input in consecutive frames.

\textit{VertexNet} is used to infer the scale information of motion. We use
vertex maps of consecutive frames as input and embed the translational motion
into the feature. \textit{NormalNet} is configured to extract the rotation
information between two frames. The output from both networks is represented as
a feature vector size of 1024, and the sum of the two feature vectors is used as input for
\textit{PoseNet}, which is designed as fully-connected networks that
transfer features for metric information, and predicts translation and rotation
separately.

\textit{VertexNet} and \textit{NormalNet} are designed based on
residual blocks \cite{he2016deep} with fully convolutional networks. For \textit{PoseNet}, we
construct the decoupled pose estimation $\mathbf{x}_{t,t+1} = [\mathbf{t}, \mathbf{q}]$ composed of translation $\mathbf{t} \in \mathbb{R}^3$ and rotation  as quaternion $\mathbf{q} \in \mathbb{R}^4$.

\subsection{Objective Losses}

In this section, we introduce two types of objective losses: unsupervised and
supervised. Both losses are selectively used according to the validity of
the training data.

\subsubsection{Unsupervised Loss}

For unsupervised training, we integrate the \ac{ICP} method into the deep-learning
framework. Given the predicted relative motion $\mathbf{x}_{t,t+1}$, we define
the orthogonal distance of the point correspondences as the loss value. For the correspondence search, projective data association is used to obtain point
correspondences. Each vertex in the vertex map $\mathbf{v}_{t+1} \in V_{t+1}$
is transformed into a frame $t$ as $\mathbf{v}'_{t} = T_{t,t+1} \mathbf{v}_{t+1}$,
where $T_{t,t+1} \in \mathbb{R}^{4\times4}$ is a transformation matrix. Next, the
corresponding vertex and normal vectors are assigned via a projection function
$\pi(\cdot)$, the mathematical expression for which is
\begin{eqnarray}
\mathbf{\bar{v}}_{t} &=& V(\pi(\mathbf{v}'_{t})) \\
\mathbf{\bar{n}}_{t} &=& N(\pi(\mathbf{v}'_{t}))
\end{eqnarray}
where $\mathbf{\bar{v}}_{t}$ and $\mathbf{\bar{n}}_{t}$ are corresponding vertex
and normal vectors of $\mathbf{v}_{t+1}$ on frame $t$, respectively. Given the point
correspondences, the ICP loss $\mathcal{L}_{icp}$ is
\begin{equation}
\mathcal{L}_{icp} = \sum_{\mathbf{v} \in V_{t+1}} \mathbf{\bar{n}}_{t} \cdot
(T_{t,t+1} \mathbf{v}_{t+1}-\mathbf{\bar{v}}_{t}),
\end{equation}
where $\mathcal{L}_{icp}$ represents the sum of the normal distances of the point
correspondences.

We also introduce field-of-view loss (FOV loss) $\mathcal{L}_{fov}$ which prevent divergence training to the out of
field-of-view condition because although the \ac{ICP} loss is essential for training convergence, additional
regularization is needed for a stable training process. Because the \ac{ICP} loss $\mathcal{L}_{icp}$ is zero when there are no correspondences, a na\"{\i}ve \ac{ICP} loss may lead the
network to a large relative motion which yields no correspondences. To avoid
such cases, we used a penalty loss as a hard-counting loss of out-of-FOV points. The FOV loss is expressed as
\begin{equation}
\mathcal{L}_{fov} = \sum_{\mathbf{v} \in V_{t+1}}
\mathbb{I}(\pi(T_{t,t+1}\mathbf{v})-(w, h)) + \mathbb{I}(-\pi(T_{t,t+1}))
\end{equation}
where $\mathbb{I}$ represents the heaviside function and $(w, h)$ are the width and height
of the vertex map, respectively. Finally, the overall unsupervised loss is obtained as
\begin{equation}
  \small
  \mathcal{L}_{uns} = \mathcal{L}_{icp}\exp(-s_{icp}) + s_{icp}
  +\mathcal{L}_{fov}\exp(-s_{fov}) + s_{fov}
  \label{eq:L_uns}
\end{equation}
where $s_{icp}$ and $s_{fov}$ are trainable scaling factors which balance the
magnitude of each loss.

\begin{figure}[!b]
	\centering
	\def\width{0.46\columnwidth}%
	\subfigure[Perturbation on translation]{%
		\includegraphics[clip, trim=0 0 0
		50, width=\width]{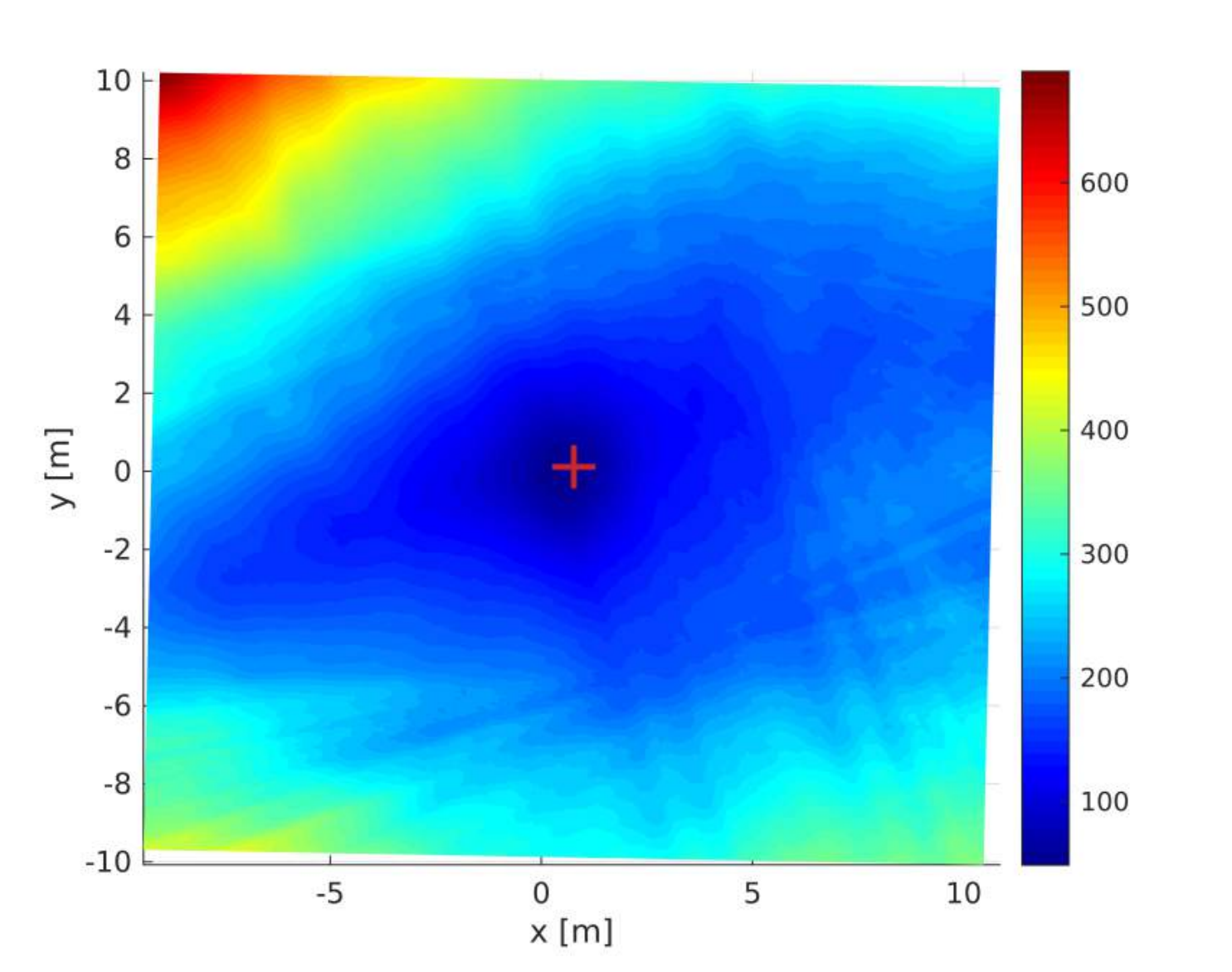}
		\label{fig:error_on_t}
	}
	\subfigure[Perturbation on rotation]{%
		\includegraphics[width=\width]{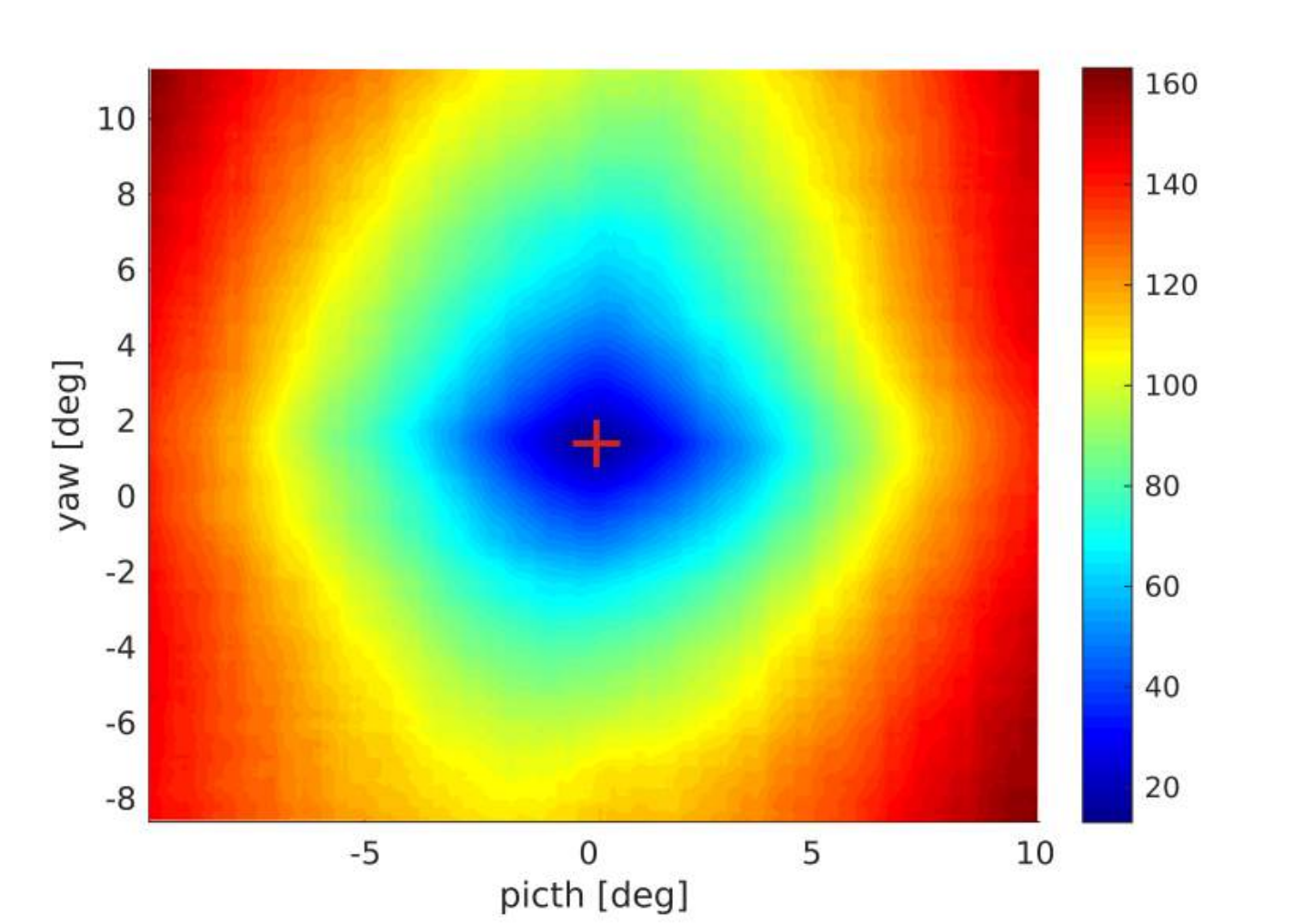}
		\label{fig:error_on_q}
	}

  \caption{Unsupervised loss ($\mathcal{L}_{uns}$) values on motion
  perturbation. To verify the unsupervised loss, we plot the tendency of the
  loss values (z-axis) over motion perturbation (x and y-axis) on the
  ground-truth pose (red cross). The figure represents the validity of the loss
  over large motion perturbation ($\pm$\unit{10}{m} on translation and $\pm
  10^\circ$ on rotation). Colors in error bars indicate the magnitude of
  unsupervised loss $\mathcal{L}_{uns}$.}

	\label{fig:loss_vals}
  \vspace{-4mm}
\end{figure}

The characteristics of loss $\mathcal{L}_{uns}$ on motion perturbation is 
depicted in \figref{fig:loss_vals}. A red cross sign ($+$) in each subplot 
is the loss value on the ground-truth relative pose. We simply add perturbation 
on
translation \figref{fig:error_on_t} and rotation \figref{fig:error_on_q}, and
track the unsupervised loss transitions on the motion errors. Each curve on translation and rotation has convex shape around ground-truth. This indicates that the tendency of loss supports the validity of the proposed loss on training.

\begin{figure}[!h]
	\centering
	\def\width{0.46\columnwidth}%
	\subfigure[Euler Loss]{%
		\includegraphics[clip, trim=110 240 120 250, width=\width]{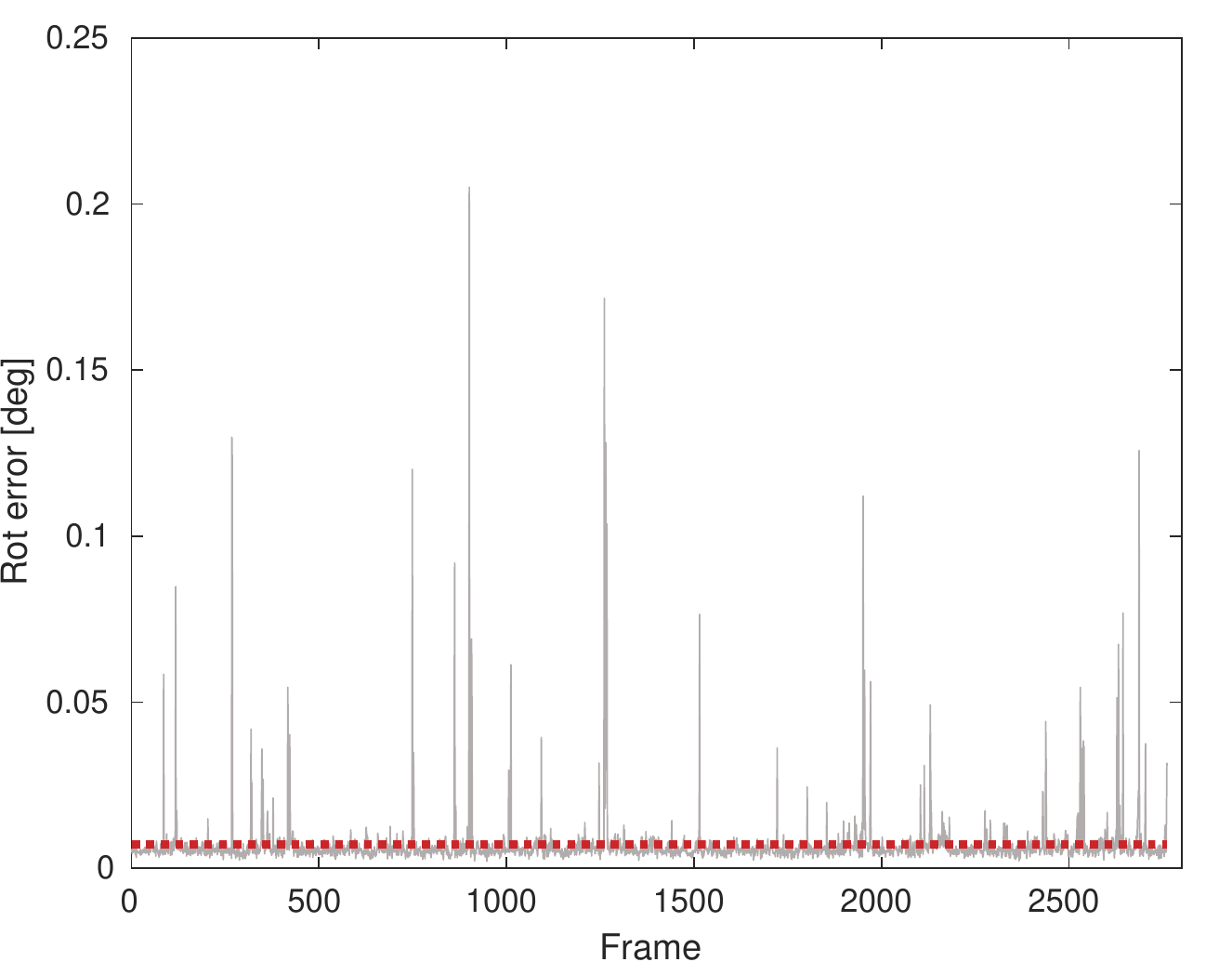}
		\label{fig:euler_on}
	}
	\subfigure[Quaternion Loss]{%
		\includegraphics[clip, trim=110 240 120 250, width=\width]{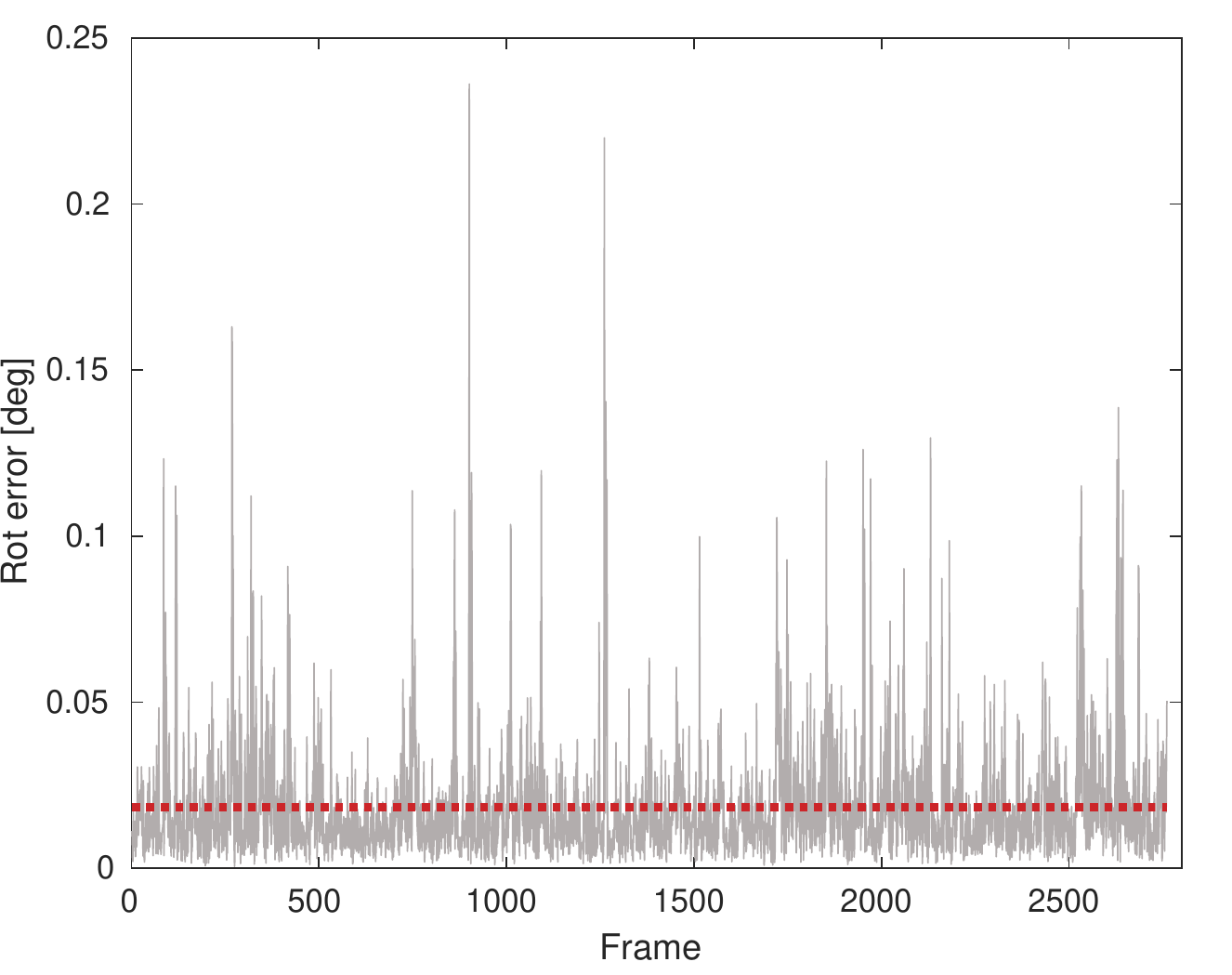}
		\label{fig:euler_off}
	}

  \caption{The rotation error of each frame. \subref{fig:euler_on} errors from
  Euler-transformed loss. \subref{fig:euler_off} errors from quaternion loss.
  The red dotted lines are the averages of the errors for all of the frames. We
  tested both losses under the same training conditions. As can be seen the
  Euler-transformed loss showed a more stable result on training compared to the
  quaternion reprsentation.}

	\label{fig:euler_mode}
  \vspace{-4mm}
\end{figure}

\subsubsection{Supervised Loss}

In this section, we describe the supervised loss that can be applied when a
ground-truth pose or a reference pose (visual odometry) is available. Similar to
\cite{kendall2015posenet,valada2018deep}, our network estimates relative
rotation as quaternion form which has bounded magnitude of rotation $[-\pi,
\pi]$. However, if quaternion subtraction is used as the training loss,
normalization is not considered and thus the rotation difference is not
reflected correctly. To overcome this issue, we transform the quaternion
$\mathbf{q}$ into the Euler angle $\mathbf{r}=[r^x, r^z, r^y]$ in degree. In
practice, we find that Euler transformed representation shows better performance
and convergence of training. Finally, the supervised loss $\mathcal{L}_{sup}$ is
expressed as follows,
\begin{equation}
\mathcal{L}_{sup} = \mathcal{L}_{t}\exp(-s_{t}) + s_{t}
+\mathcal{L}_{r}\exp(-s_{r}) + s_{r}
\label{eq:L_sup}
\end{equation}
where $\mathcal{L}_{t} = || \mathbf{t}-\mathbf{\hat{t}} ||_l$ is the translation
loss with ground-truth $\mathbf{\hat{t}}$, $\mathcal{L}_{r} = ||
\mathbf{r}-\mathbf{\hat{r}} ||_l$ is the rotation loss with ground-truth
$\mathbf{\hat{r}}$, and $s_{(\cdot)}$ represents trainable scale factors to
balance the translation and rotation losses during training. We used L1-norm as
loss values.

\figref{fig:euler_mode} clarifies the effect of the proposed rotation loss. We
verified both the Euler format \figref{fig:euler_on} and the quaternion format
\figref{fig:euler_off} with the same training setup. This plot shows the
rotation error of each frame after training had finished. As can be seen, using
the proposed representation method guides the network to learn the rotation
better.


\section{Experimental Results}
\label{experimental_results}

In this section, we evaluated the proposed supervised model (DeepLO-Sup) and unsupservised model (DeepLO-Uns) via both qualitative and quantitative comparisons using publicly available datasets, KITTI and Oxford RobotCar.

\subsection{Implementation and Training}

The proposed network was implemented using PyTorch and trained with an NVIDIA
GTX 1080ti. We employed the Adam solver \cite{kingma2014adam} with
$\beta_1=0.9$, $\beta_2=0.99$, and $w_{decay}=10^{-5}$. We started the training
with an initial learning rate of $10^{-4}$ and controlled it by a step scheduler
with a step-size of 20 and $\gamma=0.5$. The scaling factors were initialized
with $s_{(\cdot)}=-3$ for automatic scale learning on loss functions
\equref{eq:L_uns} and \equref{eq:L_sup}, and we set horizontal field of view
$f_{h} = 360\degree$ and vertical field of view $f_v = 26\degree$ to process raw
point clouds to the vertex map.  The corresponding horizontal and vertical
resolutions were  $\delta_h = 0.5\degree$ and $\delta_v=0.5\degree$, and the
size of the input vertex map was 720 $\times$ 52.

Unlike supervised learning, unsupervised learning needs guided training at the
start. The initial relative motion from the network has random values, thus we
first trained the network with fixed motion beforehand; we employed a simple
forward motion (\unit{1}{m} moving forward with no rotation) for the first 20
iterations and then switched the training to unsupervised mode.

\begin{figure*}[!t]
    \centering
    \def\width{1\textwidth}%
        \includegraphics[clip, trim=0 70 0 0,
        width=\width]{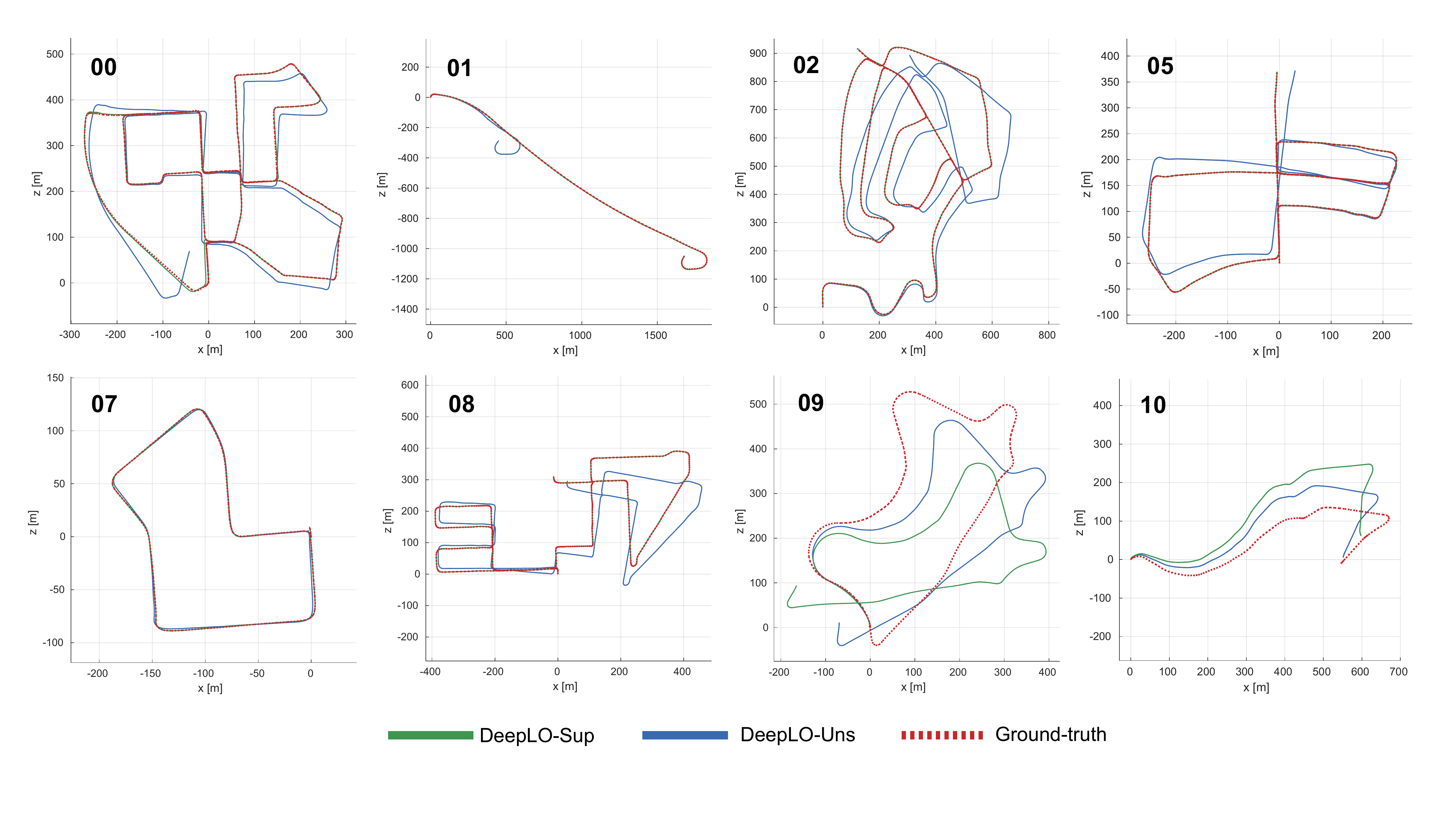}
        \label{fig:kitti_00}
    \caption{KITTI trajectory comparison of the proposed method with the
    supervised learning (DeepLO-Sup) and the unsupervised loss
    (DeepLO-Uns) and the ground-truth trajectory. As in previous
    approaches
    \cite{li2018undeepvo,zhu2018robustness,zhou2017unsupervised}, we used
    sequences \texttt{00-08} for training and \texttt{09-10} for test.
    }
    \label{fig:kitti_result}
    \vspace{-4mm}
    
\end{figure*}

\begin{table*}[t]
    \centering
    \caption{KITTI odometry evaluation.}
    \label{tab:kitti}
\begin{tabular}{cccccccccccccc}
    \hline
    & Sequence     & \multicolumn{2}{c}{0} & \multicolumn{2}{c}{1} &
    \multicolumn{2}{c}{2} & \multicolumn{2}{c}{3} & \multicolumn{2}{c}{4}  &
    \multicolumn{2}{c}{5} \\ \hline
    &              & $t_{rel}$       & $t_{rel}$     & $t_{rel}$       &
    $t_{rel}$     & $t_{rel}$
    & $t_{rel}$     & $t_{rel}$       & $t_{rel}$     & $t_{rel}$        &
    $t_{rel}$     & $t_{rel}$       &
    $t_{rel}$     \\ \hline
    \multirow{2}{*}{Proposed}       & DeepLO-Uns & 1.90       & 0.80     &
    37.83      & 0.86     & 2.05       & 0.81     & 2.85       & 1.43     &
    1.54        & 0.87     & 1.72       & 0.92     \\ \cline{2-14}
    & DeepLO-Sup   & 0.32       & 0.12     & 0.16       & 0.05     & 0.15
    & 0.05     & 0.04       & 0.01     & 0.01        & 0.01     & 0.11       &
    0.07     \\ \hline
    \multirow{3}{*}{Learning-based} & Zhu et al. \cite{zhu2018robustness}   &
    4.56       & 2.46     &
    78.98      & 3.03     & 5.89       & 2.16     & 6.84       & 2.42     &
    9.12        & 1.42     & 3.93       & 2.09     \\ \cline{2-14}
    & SfMLearner \cite{zhou2017unsupervised}  & 66.35      & 6.13     &
    35.17      & 2.74     & 58.75
    & 3.58     & 10.78      & 3.92     & 4.49        & 5.24     & 18.67      &
    4.10     \\ \cline{2-14}
    & UnDeepVO  \cite{li2018undeepvo}   & 4.41       & 1.92     & 69.07      &
    1.60     & 5.58
    & 2.44     & 5.00       & 6.17     & 4.49        & 2.13     & 3.40       &
    1.50     \\ \hline
    Model-based                     & SuMa    \cite{Behley-RSS-18}     &
    2.10       & 0.90     &
    4.00       & 1.20     & 2.30       & 0.80     & 1.40       & 0.70     &
    11.90       & 1.10     & 1.50       & 0.80     \\ \hline
    &              & \multicolumn{2}{c}{6} & \multicolumn{2}{c}{7} &
    \multicolumn{2}{c}{8} & \multicolumn{2}{c}{9} & \multicolumn{2}{c}{10}
    &            &          \\ \hline
    &              & $t_{rel}$       & $t_{rel}$     & $t_{rel}$       &
    $t_{rel}$     & $t_{rel}$
    & $t_{rel}$     & $t_{rel}$       & $t_{rel}$     & $t_{rel}$        &
    $t_{rel}$     &
    &          \\ \hline
    \multirow{2}{*}{Proposed}       & DeepLO-Uns & 0.84       & 0.47     &
    0.70       & 0.67     & 1.81       & 1.02     & 6.55       & 2.19     &
    7.74        & 2.84     &            &          \\ \cline{2-14}
    & DeepLO-Sup   & 0.03       & 0.07     & 0.08       & 0.05     & 0.09
    & 0.04     & 13.35      & 4.45     & 5.83        & 3.53     &
    &          \\ \hline
    \multirow{3}{*}{Learning-based} & Zhu et al. \cite{zhu2018robustness}  &
    7.48       & 3.76     &
    3.13       & 2.25     & 4.81       & 2.24     & 8.84       & 2.92     &
    6.65        & 3.89     &            &          \\ \cline{2-14}
    & SfMLearner \cite{zhou2017unsupervised}  & 25.88      & 4.80     &
    21.33      & 6.65     & 21.90
    & 2.91     & 18.77      & 3.21     & 14.33       & 3.30     &
    &          \\ \cline{2-14}
    & UnDeepVO  \cite{li2018undeepvo}   & 6.20       & 1.98     & 3.15       &
    2.48     & 4.08
    & 1.79     & 7.01       & 3.61     & 10.63       & 4.65     &
    &          \\ \hline
    Model-based                     & SuMa   \cite{Behley-RSS-18}      &
    1.00       & 0.60     &
    1.80       & 1.20     & 2.50       & 1.00     & 1.90       & 0.80     &
    1.80        & 1.00     &            &          \\ \hline
\end{tabular}
    \vspace{2mm}

  \caption*{Translation $t_{rel}(\%)$ and rotation $r_{rel}(\degree/100m)$ RMSE
  drift on length of \unit{100m-800m} are presented. Our model was trained on
  sequences \texttt{00-08} along with the compared methods. The RMSE values of
  the other methods were obtained from \cite{zhu2018robustness} and
  \cite{Behley-RSS-18}.}

  \vspace{-6mm}
\end{table*}

\subsection{Evaluation with the KITTI Dataset}

We evaluated our method on the well-known odometry datasets of KITTI Vision
Benchmark \cite{Geiger2012CVPR}. The KITTI dataset contains 3D point clouds from
Velodyne HDL-64E with ground-truth global 6D pose. This dataset has 10 sequences
from different environments having dynamic objects (e.g., urban, highways, and
streets).

\textbf{Training details.} Similar to previously reported learning-based
odometry methods \cite{li2018undeepvo,zhu2018robustness}, we used sequences
\texttt{00-08} for the training and \texttt{09-10} for the test. In addition, we
verified our method via both training strategies (i.e., the supervised and the
unsupervised).

\textbf{Evaluation.} \figref{fig:kitti_result} shows the trajectory comparisons
of the proposed methods with different strategies. The performance of the
trajectories represents the soundness of the network fit (training:
\texttt{00}-\texttt{08}) and the generality of our method (test:
\texttt{09}-\texttt{10}). Note that all of the tests had the same parameter
settings on both learning models: DeepLO-Sup and DeepLO-Uns.

The trajectories from the training sets (\texttt{00} to \texttt{08}) of
DeepLO-Sup showed well-fitted result to ground-truth, but the performance was
relatively low with the test sequences. However, with DeepLO-Uns, the results
with both the training and test sequences conveyed similar performance. This
finding indicates that unsupervised learning attained a better performance for
the generality aspect.

\tabref{tab:kitti} contains the details of the results; the average translation
$t_{rel}(\%)$ and rotation $r_{rel}(\degree/100m)$ RMSE drift on length of
\unit{100-800}{m}. We evaluated our method against several previously reported
ones, namely UndeepVO \cite{li2018undeepvo}, SfMLeaner
\cite{zhou2017unsupervised}, and Zhu's method \cite{zhu2018robustness}; the
result values for these were taken from the result in \cite{zhu2018robustness}.
We also compared our method to SuMa \cite{Behley-RSS-18} which is recent
model-based \ac{SLAM} using LiDAR measurements. The values of SuMa are
referenced from frame-to-frame estimation results.

Compared to learning-based methods, we can see that our method gives better
results for learning data sets. Unlike DeepLO-Sup, however, DeepLO-Uns showed a
large translation error in sequence \texttt{01}. Sequence \texttt{01} (highway)
includes has dynamic objects, fewer structures and large translational motions
than other sequences. Thus, it is difficult to capture true translation with the
unsupervised loss which relies on geometric consistency of structures. This
aspect also can be seen in other learning-based methods. For the test sequences,
DeepLO-Uns performed better than other learning-based methods when DeepLO-Sup
was slightly worse than DeepLO-Uns for the test sequences. This is because
DeepLO-Sup is overfitted to the training sets, and it is expected that better
results will be obtained by adding more validation sets to prevent overfitting.
Compared to SuMa, DeepLo-Uns showed excellent performance for learning data
because it repeatedly learns the geometric and motion characteristics of the
dataset. To achieve better performance in test sequences, we could train more
sequences with various motions and guide the training loss with robust kernels
which reduce the effects of dynamic objects.

\begin{figure*}[!t]
    \centering
    \includegraphics[clip, trim=0 0 0 0,
    width=1\textwidth]{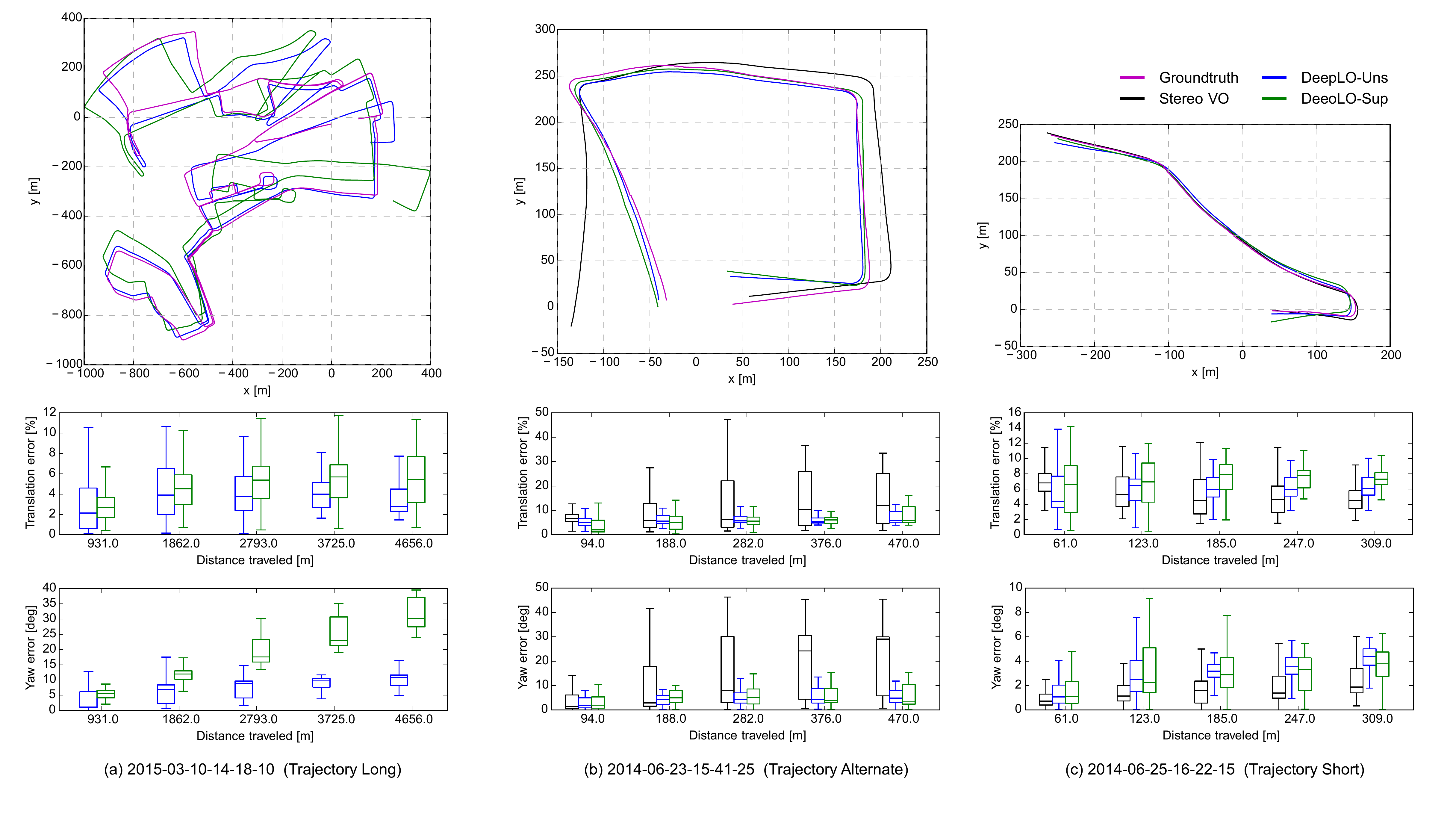}%
    \caption{Evaluation of the proposed methods via the Oxford RobotCar
    dataset; trajectories (first row), relative translation errors over
    distance (second
        row),
        and relative heading errors over distance (third row). Each column
        shows the evaluation with diffent type of sequences. (a) Training:
        \texttt{2014-03-10-14-18-10} \texttt{Long}, (b) Test:
        \texttt{2014-06-23-15-41-25}
        \texttt{Alternate}, and (c) Test: \texttt{2014-06-25-16-22-15}
        \texttt{Short}.
        Box plots represent the error statistics: the median (center line),
        25\% and 75\% quantiles (box), and minimum and maximum errors (whisker).
    }

    \label{fig:oxford_result}
    \vspace{-4mm}
\end{figure*}

\subsection{Evaluation with the Oxford RobotCar Dataset}

The Oxford RobotCar dataset \cite{maddern20171} comprises data collected by
repeating the same path dozens of times for long-term autonomy researches.

\textbf{Dataset preparation.} Because this dataset uses a push-broom style 2D
LiDAR, information from a single scan is not enough to train a network and infer
the robot pose. Therefore, we made a submap with sufficient length (\unit{80}{m}
in our work) of accumulated 2D scans using each scan pose interpolated from the
\ac{INS} data. We determined that the ground-truth pose corresponding to the
submap was the interpolated global pose (in the world frame) of a center scan.
The 3D point coordinates of the submap are represented in the robot frame where
the ground-truth global pose of the submap was considered as the origin. Each
submap as a 3D scan is sampled per every \unit{1-2m} using a truncated normal
distribution.

\textbf{Training details.} The Oxford Robotcar dataset can be divided into three
types: \texttt{Long}, \texttt{Alternate} and \texttt{Short}. Since the focus of
the Oxford RobotCar dataset is on seasonal diversity, we used \texttt{Long}
sequences (\texttt{2015-02-03-08-45-10} and \texttt{2015-03-10-14-18-10}) as
training data and the other two types (\texttt{Alternate} and \texttt{Short}) as
test data. It is notable that the Oxford RobotCar dataset has less trajectory
diversity than the KITTI dataset, thus we applied transfer learning to secure
network generality and improve learning stability, and we used the weight of the
network learned by the KITTI dataset as the initial weight. We found that this
strategy significantly improved the training phase by enabling the learning for
the dataset with less diversity for training.

\textbf{Evaluation.} We compared our method with the ground-truth trajectory and
stereo visual odometry given in the Oxford RobotCar dataset. Since all
trajectories have different framerate, we evaluated each method using the
trajectory evaluation method \cite{zhang2018tutorial}.

To compute relative errors, sub-trajectory segments of tested methods are
selected along different travel distances. Each sub-trajectory is aligned using
the first state, and the error are calculated for all the sub-trajectories.
\figref{fig:oxford_result} shows the aligned trajectories using the entire
trajectory statement. The figure includes the trajectory comparisons to the
ground-truth and corresponding error plots (relative translation error (\%) and
heading error (deg)).

For trajectory alignment, $SE(3)$ transformation was estimated and applied to
the tested methods. Each trajectory in \figref{fig:oxford_result} represents the
results on the learned sequence (\figref{fig:oxford_result}(a)) and test
sequences (\figref{fig:oxford_result}(b) and (c)), and below each trajectory is
the quantitative result compared to the ground-truth. The trajectories might be
transformed from the initial position due to the alignment. We compared our
methods with the stereo odometry provided by the dataset.

\begin{table}[]
    \caption{Trajectory Errors on Oxford RobotCar dataset}
    \label{tab:oxford}
    \begin{tabular}{cc|ccc}
        \hline
        &                                                             &
        \multicolumn{3}{c}{Absolute Trajectory Error (RMSE)} \\ \hline
        Datetime                                                       &
        Type                                                        &
        StereoVO  \cite{maddern20171}     & DeepLO-S       & DeepLO-U
        \\ \hline
        \begin{tabular}[c]{@{}c@{}}2014-05-14\\ -13-50-20\end{tabular} &
        Alternate                                                   &
        37.74          & 14.71            & 19.93            \\ \hline
        \begin{tabular}[c]{@{}c@{}}2014-05-14\\ -13-59-05\end{tabular} &
        \begin{tabular}[c]{@{}c@{}}Alternate\\ Reverse\end{tabular} &
        34.55          & 12.89            & 22.93            \\ \hline
        \begin{tabular}[c]{@{}c@{}}2014-06-23\\ -15-41-25\end{tabular} &
        Alternate                                                   &
        36.09          & 16.94            & 12.80            \\ \hline
        \begin{tabular}[c]{@{}c@{}}2014-06-25\\ -16-22-15\end{tabular} &
        Short                                                       &
        4.22           & 9.53             & 6.78             \\ \hline
    \end{tabular}
    \vspace{-4mm}
\end{table}

In the case of the training sequence, the stereo trajectory was not accurate
enough to be used as the baseline and was thus excluded from the comparison. As
can be seen, our methods showed stable and comparable performances for all of
the sequences with the trained networks being able to capture the relative
motion of test sets. Note that our proposed models achieved performances that
were better or close to StereoVO \cite{maddern20171}.

\tabref{tab:oxford} represents absolute trajectory errors (ATE) on test
sequences. We compared translation error (m) of all trajectories. First two
sequences of the table are \texttt{Alternate} and \texttt{Alternate} with
reverse direction which is not introduced in \figref{fig:oxford_result}. This
evaluation quantifies the quality of the whole trajectory. Since ATE is
sensitive to the initial results on the path, interesting results can be seen
compared to relative errors. Looking at the results for sequence
\texttt{2014-06-25-16-22-15}, we found that the error at the beginning of
DeepLO-Uns and DeepLO-Sup also affected the absolute trajectory.

\section{Conclusion}
\label{conclusion}

We demonstrated a novel learning-based LiDAR odometry estimation
pipeline in an unsupervised and a supervised manner. We suggested surfel-like
representations (vertex and normal map) as network inputs without precision loss.
We showed that the proposed unsupervised loss could capture the geometric
consistency of point clouds. To the best of our knowledge, ours is the first unsupervised approach for deep-learning-based LiDAR odometry. In addition, our method showed prominent
performance compared to other learning-based or model-based methods in various
environments. We also derived training adaptation via transfer learning in heterogeneous environments. In future work, we plan to design the networks and
loss functions for large and fast motion such as in sequence \texttt{01} of the KITTI
dataset and extend our framework to sequential approaches with recurrent
neural networks.


\balance
\small\bibliographystyle{IEEEtranN} 
\bibliography{string-short,references}

\end{document}